\documentclass[runningheads]{llncs}

\usepackage[mobile]{eccv} % arXiv

\usepackage{eccvabbrv}

\usepackage{graphicx}

\usepackage{lmodern}

\usepackage{algorithm}
\usepackage{algpseudocode}
\usepackage{amsfonts}
\usepackage{amsmath}
\usepackage{amssymb}
\usepackage{booktabs}
\usepackage{color}
\usepackage{letltxmacro}
\usepackage{tabularray}

\usepackage[accsupp]{axessibility}

\usepackage[hidelinks]{hyperref}

\usepackage{orcidlink}

\UseTblrLibrary{booktabs}

\title{GryphOne: Symbol-Aware Masked Diffusion for Structural Refinement in Offline Handwritten Mathematical Expression Recognition}

\titlerunning{GryphOne}

\author{Takaya Kawakatsu\inst{1}\orcidlink{0000-0003-1285-2748} \and Ryo Ishiyama\inst{2}\orcidlink{0009-0007-0162-9950}}

\authorrunning{T. Kawakatsu and R. Ishiyama}

\institute{Preferred Networks, Inc., Otemachi, Tokyo, Japan\\
\email{kat.nii.ac.jp@gmail.com}\\
\url{https://researchmap.jp/t.kat}
\and
Kyushu University, Fukuoka, Japan\\
{\tt\small ryo.ishiyama@human.ait.kyushu-u.ac.jp}\\
\url{https://researchmap.jp/rishiyama}}

\LetLtxMacro\oldeqref\eqref
\renewcommand{\eqref}[1]{Eq.~\oldeqref{eq:#1}}

\NewDocumentCommand\secref{m}{Section~\ref{sec:#1}}
\NewDocumentCommand\figref{m}{Fig.~\ref{fig:#1}}
\NewDocumentCommand\tabref{m}{Table~\ref{tab:#1}}
\NewDocumentCommand\tabsref{mm}{Table~\ref{tab:#1}--\ref{tab:#2}}
\RenewDocumentCommand\algref{m}{Alg.~\ref{alg:#1}}

\NewDocumentCommand\CM{}{\checkmark}

\NewDocumentCommand{\CER}{}{CER $\downarrow$}
\NewDocumentCommand{\EM}{}{EM $\uparrow$}
\NewDocumentCommand{\ER}{}{ExpRate $\uparrow$}
\NewDocumentCommand{\ERone}{}{$\leq$ 1 $\uparrow$}

\NewDocumentCommand{\SER}{}{SER $\downarrow$}
\NewDocumentCommand{\FPS}{}{FPS $\uparrow$}

\begin{document}
\maketitle

\begin{abstract}
Handwritten mathematical expression recognition (HMER) requires reasoning over diverse symbols and structures, yet autoregressive models struggle with exposure bias and syntax inconsistency.
We present GryphOne, a discrete diffusion framework which reformulates HMER as iterative symbolic refinement instead of sequential generation.
GryphOne progressively refines symbols and relations, removing autoregression and improving consistency.
Symbol-aware tokenization and random-masking mutual learning further enhance robustness to handwriting diversity.
On the MathWriting benchmark, GryphOne achieves 5.51\% CER and 59.9\% EM (ExpRate), outperforming all reimplemented models in the matched setting as well as the commercial HMER system.
Held-out evaluation on CROHME 2014--2023 further shows strong cross-dataset generalization.
\keywords{OCR \and Handwritten \and Math \and Masked Diffusion Model}
\end{abstract}

\section{Introduction}

\begin{figure}[tb]
\centering
\includegraphics[width=\textwidth]{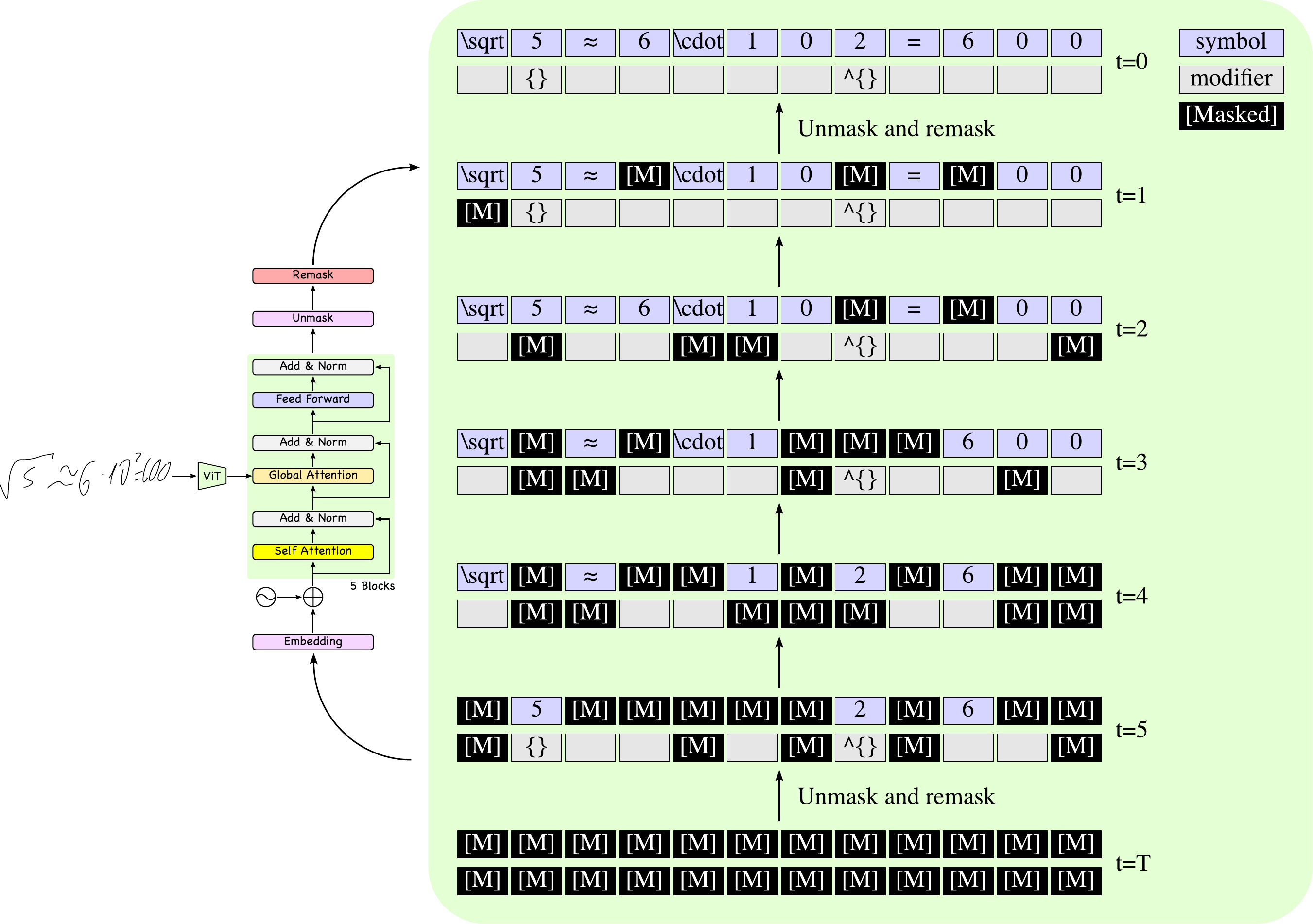}
\caption{Illustration of the symbolic reverse diffusion process in GryphOne.}
\label{fig:key}
\end{figure}

Handwritten mathematical expression recognition (HMER) aims to understand mathematical symbols and convert them into \LaTeX{}.  
HMER is more challenging than general optical character recognition (OCR) due to two types of ambiguity:  
(1) \textit{symbolic ambiguity}, caused by diverse writing styles and symbol shapes; and  
(2) \textit{syntactic ambiguity}, arising from symbol layouts.
For example, a cursive ``z'' may resemble ``2'', and $\frac{a+b}{c}$ can be misread as $a+\frac{b}{c}$.
HMER thus requires both local visual perception and globally consistent structural reasoning.

Most prior approaches~\cite{BTTR21,COMER22,ICAL24,TAMER25} adopt an autoregressive (AR) encoder--decoder framework, which generates \LaTeX{} tokens sequentially but suffers from exposure bias, where early prediction errors propagate through autoregression.
In complex formulas, the attention mechanism often fails to capture symbolic and syntactic cues, leading to over- or under-parsing.
As \LaTeX{} syntax distributes dependencies across many interrelated tokens, even a small mistake can cascade and is difficult to correct locally.

To overcome these limitations, syntax-aware models~\cite{SAN22, CAN22, TAMER25, Pos25} encode syntax trees or graphs for syntax consistency, while non-AR (NAR) models~\cite{NAMER25} mitigate exposure bias via symbol detection and arrangement.
Syntax-aware models often lack grammatical flexibility, and NAR models struggle to refine incorrect symbols or structures.  
An ideal model should unify structural consistency and robustness to ambiguity.

We present GryphOne, a masked diffusion model (MDM) which reformulates HMER as an \textit{iterative symbolic refinement} process.  
Instead of generating tokens sequentially, it operates on a fixed-length masked sequence.
At each iteration, a subset of tokens is randomly unmasked, predicted, and then remasked, gradually reconstructing the expression.
This formulation decreases exposure bias inherent in AR decoding while enabling gradual refinement of both symbols and structural relations—capabilities beyond standard NAR models.
\figref{key} shows this key idea.

To the best of our knowledge, this work is the first to apply masked diffusion to HMER.
In \LaTeX{}, hierarchical symbol relations are explicitly encoded through structural tokens such as braces, superscripts, and subscripts.
Even a trivial edit may require inserting or deleting many dependent tokens.
Such operations violate length consistency and the locality assumption underlying masked diffusion.

To enhance symbolic and syntactic refinement, we introduce a symbol-aware tokenization (SAT) scheme which decomposes \LaTeX{} tokens into visible symbols and invisible tokens in a one-to-one correspondence (\figref{SAT}).
This representation preserves syntax integrity while simplifying grammatical dependencies, allowing local symbol refinements without breaking global structure.  
We further introduce a random-masking mutual learning (RMML) method, which enforces consistency across different masks by minimizing the KL divergence~\cite{KL51}, improving robustness to handwriting variation and structural ambiguity.

Comprehensive experiments on MathWriting~\cite{MATH25} demonstrate that GryphOne outperforms all reimplemented baselines under the same training and evaluation setup, achieving the lowest character error rate (CER) and the highest expression recognition rate (ExpRate).
Held-out evaluation on CROHME~\cite{CROHME14,CROHME16,CROHME19,CROHME23} further demonstrates strong cross-dataset generalization, supporting the effectiveness of iterative symbolic refinement beyond MathWriting.

\textbf{Contributions.}
\begin{itemize}
\item Formulate HMER as a masked diffusion process, enabling iterative symbolic refinement without exposure bias.
\item Propose a SAT scheme that preserves syntactic consistency while decreasing representational complexity.
\item Design RMML to improve robustness to handwriting diversity and structural ambiguity.
\item Empirically validate the proposal on MathWriting and CROHME, achieving the best recognition performance and improved structural stability.
\end{itemize}

\section{Related Work}

Offline HMER aims to convert handwritten mathematical expression images into structured markup representations such as \LaTeX.
HMER requires understanding complex spatial layouts and hierarchical relations among visible symbols such as superscripts, subscripts, fractions, and radicals.
CROHME~\cite{CROHME14,CROHME16,CROHME19,CROHME23} has played an important role in standardizing this task, providing datasets (2014--2023) that remain the de facto benchmarks for academic comparison.

HMER has evolved through three major stages: a transition from rule-based methods to deep AR models, the emergence of syntax-aware decoders, and recent expansions toward large-scale learning, non-autoregressive (NAR) inference, and multimodal reasoning.
We summarize each direction below.

\subsection{AR Models}

HMER has been formulated as an image-to-language translation task.
BTTR~\cite{BTTR21} achieved bidirectional training within a Transformer~\cite{Tr17} decoder, improving the balance of contextual dependencies.
ABM~\cite{ABM21} introduced two decoders operating in opposite directions and performing deep mutual learning~\cite{DML18} to capture both local and global contexts.
CoMER~\cite{COMER22} introduced an explicit coverage modeling module to mitigate over-parsing and under-parsing errors.
GCN~\cite{GCN23} introduced an auxiliary prediction subtask for symbol categories (e.g., digits, operators, and relations), thereby reducing symbolic ambiguity.
ICAL~\cite{ICAL24} incorporated implicit character-aided learning, predicting unobserved structural markers to strengthen contextual reasoning and syntax recovery.
These advances improved token-level accuracy but still suffered from exposure bias due to left-to-right decoding.

\subsection{NAR Models}

NAMER~\cite{NAMER25} introduced the first NAR framework.
It consists of a visually aware tokenizer and a graph decoder that jointly predict tokens and relations, resulting in faster inference.
While NAR inference alleviates exposure bias, NAMER does not support iterative refinement of ambiguous symbols and structures.
Our work builds on this direction by introducing a discrete diffusion process which enables symbolic and structural refinement, thereby ensuring syntax consistency beyond one-shot NAR inference.

\subsection{Syntax Models}

Given that expressions inherently form hierarchical tree structures, several works have explored structure-aware modeling.
DWAP-TD~\cite{DWAPTD20} generated hierarchical trees directly instead of linear token sequences, achieving effective generalization on deeply nested expressions.
TDv2~\cite{TDv2} removed fixed traversal order constraints for more flexible generation.
SAN~\cite{SAN22} embedded grammar rules and syntax-tree guidance to reduce errors, while CAN~\cite{CAN22} introduced a symbol-counting subtask to prevent symbol omissions and duplications.
TAMER~\cite{TAMER25} evaluated structural validity during inference.
PosFormer~\cite{Pos25} incorporated a relative positioning task that stacks vertical relations (e.g., superscript and subscript) and decodes them during training.
These approaches still depend on AR inference, which can limit opportunities for global structural refinement at inference time.

\subsection{VLMs}

Recent studies integrate vision and language modeling (VLM) to further improve generalization.
For example, Uni-MuMER~\cite{UNIMUMER25} trains large VLMs under a unified multi-task objective, which incorporates syntax-aware chain-of-thought and error correction, thereby demonstrating the strong potential of VLM.
However, it still relies on AR inference and does not explicitly address structural ambiguity.
Our work employs symbol-aware diffusion to unify symbolic and syntactic refinement in a non-generative recognition setting.

\section{Methodology}

GryphOne departs fundamentally from AR architectures.
Instead of sequentially generating \LaTeX{} tokens, we formulate HMER as an \emph{iterative refinement} process within a MDM.
\figref{key} presents an overview of the proposed architecture.
Starting from a fully masked symbol sequence, the model progressively refines all symbols in parallel, eliminating causal dependencies and mitigating exposure bias.

\subsection{Overall Pipeline}

GryphOne consists of a vision Transformer (ViT) encoder and a masked-diffusion decoder that operates on masked symbol sequences.

\subsubsection{Input}

The handwritten strokes are rasterized into a static image, which is then encoded by a ViT to extract visual features.
The corresponding \LaTeX{} expression is converted into a symbol sequence using SAT introduced in \secref{SAT}, which aligns each visible symbol with its associated structural modifiers.

\subsubsection{Training}

During training, the model learns to denoise partially masked symbol sequences.
For each sequence, a continuous diffusion time $t\sim\mathcal{U}(0,T)$ is sampled at random.
The sequence $x_0$ is then corrupted into a partially masked sequence $x_t$ using the forward diffusion process shown in \algref{forward}.
The decoder predicts the original symbols in a single refinement step, optimized using a cross-entropy loss over all sequence positions, including both masked and unmasked positions.

\subsubsection{Inference}

At inference time, decoding starts from a fully masked sequence and iteratively refines the sequence over $T$ steps via random remasking and prediction shown in \algref{reverse}, yielding the final $x_0$.

\subsection{Masked Diffusion Formulation}

Given a ground-truth sequence $x_0$, we define both \textit{forward} and \textit{reverse} diffusion processes.
\algref{forward} and \algref{reverse} describe the two processes, respectively.
The former is applied during training and reused for remasking at inference time, while the latter is applied during inference.

\subsubsection{Forward Diffusion Process}

Each symbol in $x_0$ is independently replaced with a \texttt{MASK} token with probability $\frac{t}{T}$, yielding a masked sequence $x_t$.
$t$ is a diffusion timestep sampled at random, and $T$ denotes the number of diffusion steps.
This forward diffusion process exposes the model to diverse masking levels and enables the model to learn to unmask (denoise) $x_t$ from arbitrary partial observations.

\subsubsection{Reverse Diffusion Process}

Inference starts from a masked sequence $x_T$ with fixed length.
The complete sequence $x_0$ is obtained by repeating the refinement process for $T$ steps.
Here, we discretize the diffusion time $t$ into $T$ steps.
At each refinement step $t$, the decoder predicts, in parallel, the target symbols at masked positions and then remasks a subset of tokens according to the forward diffusion process at time $t$.
By alternating between unmasking and remasking, the model is encouraged to iteratively refine predictions under diverse masking patterns.

\begin{algorithm}[t]
\caption{Forward Diffusion (Random Masking) Process}
\label{alg:forward}
\begin{algorithmic}[1]
\State \textbf{Input:} Ground-truth sequence $x_0$ of length $L$, diffusion time $t$, total steps $T$
\For{$i = 1 \dots L$}
\State Sample $u_i \sim \mathcal{U}(0,1)$
\If{$u_i < \frac{t}{T}$}
\State $x_{t,i} \leftarrow \texttt{MASK}$
\Else
\State $x_{t,i} \leftarrow x_{0,i}$
\EndIf
\EndFor
\State \textbf{Output:} Partially masked sequence $x_t$
\end{algorithmic}
\end{algorithm}

\begin{algorithm}[t]
\caption{Reverse Diffusion (Iterative Refinement) Process}
\label{alg:reverse}
\begin{algorithmic}[1]
\State \textbf{Initialize:} $x_T \gets [\texttt{MASK}, \dots, \texttt{MASK}]$
\For{$t = T, \dots, 1$}
\State Predict symbol distributions $p(x_0 \mid x_t)$ in parallel
\State $\hat{x}_0 \gets \arg\max p(x_0 \mid x_t)$
\State $x_{t-1} \gets \text{Random-Remask}(\hat{x}_0, t-1)$
\EndFor
\State \textbf{Output:} Refined symbol sequence $x_0$
\end{algorithmic}
\end{algorithm}

\subsubsection{Effect}

The diffusion formulation removes left-to-right causal dependencies and refines all symbols in parallel, mitigating exposure bias while enabling repeated global correction.
Stochastic corruption at each diffusion step prevents premature convergence to locally plausible but globally inconsistent structures.
As observed in \secref{epoch}, this is associated with gradual error decay and stable refinement dynamics.

\subsection{Symbol-Aware Tokenization (SAT)\label{sec:SAT}}

\begin{figure}[tb]
\centering
\includegraphics[width=\textwidth]{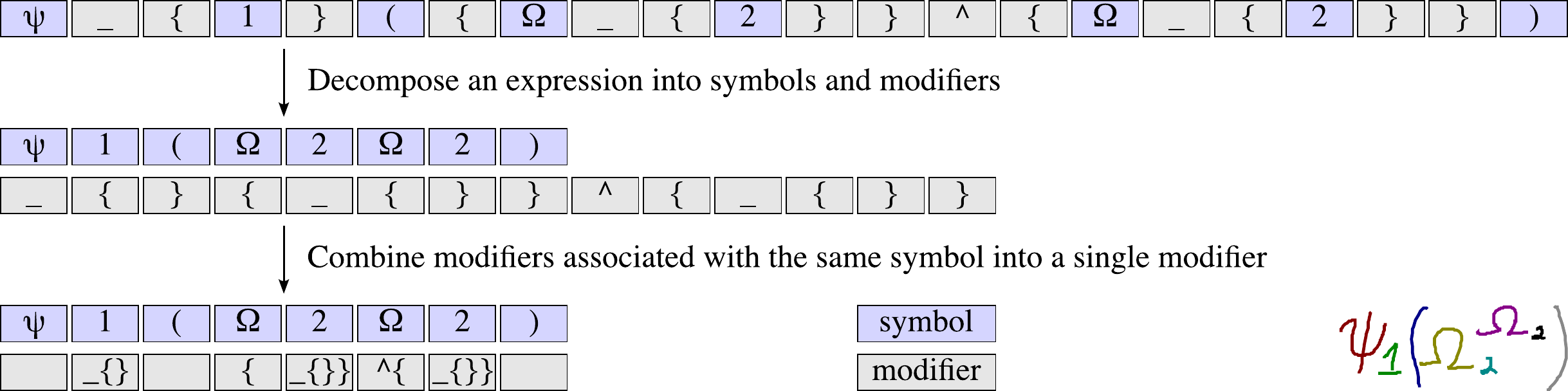}
\caption{Symbol-aware tokenization (SAT).}
\label{fig:SAT}
\end{figure}

Applying diffusion directly to raw \LaTeX{} sequences is unstable, as local structural edits may affect multiple dependent tokens and break syntactic consistency.
For example, removing a superscript in ${x_{1}}^{y_{2}}$ to obtain $x_{1}y_{2}$ eliminates five tokens in total, including a caret and four bracket tokens, illustrating how a local structural change can induce large, non-local modifications in the token sequence.
To enable locality-preserving refinement, we propose SAT as illustrated in \figref{SAT}.

\subsubsection{Method}

SAT decomposes a \LaTeX{} expression into a sequence of visible symbols and a sequence of invisible modifiers.
Visible symbols include digits, letters, and operators, while modifiers include carets, underscores, and braces.
Modifiers are aligned to the positions of their corresponding symbols by inserting empty tokens when necessary.
Carets, underscores, and open braces are aligned to the leftmost symbol appearing to their right, whereas close braces are aligned to the rightmost symbol appearing to their left.
Modifiers associated with the same symbol token are combined and assigned as a single modifier token.
The final embeddings are formed by element-wise summation of the symbol embeddings and their aligned modifier embeddings.
Each summed embedding has a one-to-one correspondence with a handwritten glyph, as defined in \eqref{embed} for each symbol token $x_{t,i}$.
\begin{equation}
\label{eq:embed}
\mathbf{E}_\text{SAT}(x_{t,i}) =
\mathbf{E}_\text{sym}(x_{t,i}) +
\mathbf{E}_\text{mod}(x_{t,i}).
\end{equation}

\subsubsection{Effect}

SAT defines diffusion units as symbol--modifier pairs aligned to individual glyphs, preserving sequence length under local structural edits.
Unlike raw \LaTeX{} tokens, where a single structural change may affect multiple tokens, SAT confines each edit to a single aligned token.
This enforces locality in the diffusion process and accelerates early-stage refinement, as observed in the reduced error rates in the initial diffusion steps in \secref{epoch}.

\subsection{Random-Masking Mutual Learning (RMML)}

The stochasticity of masked diffusion models during both training and inference can lead to inconsistent predictions across different masking patterns.
To improve stability and robustness, we introduce RMML during training as a regularization strategy.

\subsubsection{Method}

RMML generates two independently masked sequences $x'_t$ and $x''_t$ from a ground-truth $x_0$ within a single training iteration.
Both sequences are processed by the same decoder with shared parameters.
For each view, a cross-entropy loss is computed against $x_0$, and a Kullback--Leibler divergence is computed between the two predictions $\hat{x'_t}$ and $\hat{x''_t}$.
No temperature scaling is applied.
\eqref{KL} defines the loss over the entire sequence, including both masked and unmasked tokens.
\begin{equation}
\label{eq:KL}
\mathcal{L} =
\text{CE}(\hat{x'_t},x_0) +
\text{CE}(\hat{x''_t},x_0) +
D_\text{KL}(\hat{x'_t} \| \hat{x''_t}) +
D_\text{KL}(\hat{x''_t} \| \hat{x'_t}).
\end{equation}

\subsubsection{Effect}

By encouraging invariance to random masking patterns, RMML reduces sensitivity to specific noise realizations.
This results in more stable and consistent refinement trajectories at inference time.
RMML is applied only during training and does not introduce additional computational overhead during inference.

\subsection{Diffusion vs Confidence-Guided Masking}

GryphOne resembles iterative masked modeling~\cite{Mask19} in its parallel prediction over masked tokens, but employs schedule-driven stochastic remasking.
\eqref{forward} defines the forward corruption process that independently masks each token in $x_0$ with probability $\beta_t = t/T$.
\begin{equation}
\label{eq:forward}
q(x_t \mid x_0)=\prod_{i=1}^{L}\Big[(1-\beta_t)\delta(x_{t,i}=x_{0,i})+\beta_t\delta(x_{t,i}=\texttt{MASK})\Big].
\end{equation}

The model learns to reverse this process by recovering original symbols from masked inputs.
The same schedule is used at inference, so decoding mirrors the training-time reconstruction process.
In contrast, iterative masked modeling uses confidence-based remasking without an explicit corruption model.
Its refinement is not grounded in a fixed stochastic denoising schedule.
GryphOne thus adopts a discrete diffusion framework with explicit forward and reverse processes, rather than a heuristic masking strategy.

Under this absorbing corruption process, the reverse process learns to restore masked positions to their original clean tokens.
As a result, the discrete diffusion objective is formulated as denoising under a fixed categorical corruption process and is therefore not equivalent to generic masked language modeling.

\subsection{Limitations}

GryphOne has three major limitations.
Addressing them through more efficient diffusion algorithms, hybrid training schemes, or adaptive-resolution encoders is an important direction for future work.

\subsubsection{Latency}

The iterative refinement process requires multiple decoder passes and is therefore slower than one-shot NAR inference.
The number of refinement steps is often much smaller than the sequence length in practice, and each step is fully parallel over token positions.
As a result, GryphOne is faster than AR baselines in our experiments.

\subsubsection{Dataset}

Due to the data-hungry nature of diffusion-based training, GryphOne requires a large-scale and diverse training set, such as MathWriting~\cite{MATH25}, to achieve competitive reasoning performance.
Consequently, training solely on small-scale datasets such as CROHME~\cite{CROHME14,CROHME16,CROHME19} is insufficient.

\subsubsection{Input}

The current implementation relies on a ViT operating on fixed-resolution rasterized images, which may limit flexibility when handling inputs with extreme aspect ratios or very long expressions.

\section{Experiments\label{sec:exp}}

We conduct extensive experiments to evaluate the effectiveness of GryphOne on standard HMER benchmarks.

\subsection{Implementation Details}

GryphOne is implemented in PyTorch, and the encoder employs DINO~\cite{DINO21} with a patch size of 8 and a hidden dimension of 384.
The encoder operates on images with a resolution of 224 by 224 pixels.
The decoder consists of five Transformer blocks.
All attention layers use 8 heads with 30\% dropout.
Expression sequences are padded to a maximum length of 150.
The number $T$ of diffusion steps is set to 50 unless otherwise specified.
The model is trained for 60 epochs with a batch size of 32 using AdamW~\cite{ADAMW19} with a learning rate of $10^{-4}$ and a weight decay of $10^{-3}$.
Inference speed (FPS) is measured using a single NVIDIA V100 GPU.
All results are averaged over 10 inference runs.
No data augmentation is applied.

\subsection{Datasets}

We evaluate GryphOne on two standard benchmarks for HMER: MathWriting~\cite{MATH25} and CROHME~\cite{CROHME14,CROHME16,CROHME19,CROHME23}.
These datasets are provided in InkML format, which represents handwritten expressions as vector graphics, along with corresponding \LaTeX{} annotations.
Each sample is rasterized into a fixed-resolution image of 224 by 224 pixels with a 1-pixel stroke width in our experiments.

\subsubsection{MathWriting}

This set contains 230k handwritten samples and 400k synthetic samples.
The latter are synthesized by fitting handwritten symbols into compiled \LaTeX{} formulas, resulting in expressions with higher structural complexity which improve generalization.
The dataset provides standardized expressions, enabling text-based evaluation.

\subsubsection{CROHME}

This set is the most widely used benchmark for HMER.
We use the 2014, 2016, 2019, and 2023 editions for model evaluation.
The \LaTeX{} expressions are not standardized, requiring a graph-based evaluation pipeline~\cite{CROHME13} to handle omitted braces and unordered superscripts and subscripts.

\subsection{Baselines}

We used representative baselines~\cite{BTTR21,COMER22,ICAL24,TAMER25,Pos25} for which public implementations are available.
Due to the limited size of CROHME, all baselines and GryphOne are trained on MathWriting, while CROHME is used as a held-out benchmark.
This cross-dataset setting enables fair comparison under identical training conditions.

\subsection{Metrics}

We evaluate recognition performance following the standard evaluation protocols of MathWriting~\cite{MATH25} and CROHME~\cite{CROHME14,CROHME16,CROHME19,CROHME23}.

\subsubsection{MathWriting}

Character error rate (CER) and exact match (EM) are computed following the official protocol.
CER measures the ratio of incorrect tokens to the total number of tokens.
EM is equivalent to ExpRate and denotes the proportion of perfectly recognized expressions.
We also report the proportion of expressions with at most one token error.
In addition, syntax error rate (SER) is defined for ablation studies as the proportion of predictions with unmatched braces.

\subsubsection{CROHME}

ExpRate (EM) is reported using the official toolkit, which performs graph-based equivalence evaluation.
This accounts for mathematically equivalent expressions even when token order differs and constitutes the standard evaluation protocol for CROHME.

\begin{table}[tb]
\centering
\caption{MathWriting results without data augmentation.}
\label{tab:math}
\begin{tblr}{
colspec={lccccccc},
cell{1}{2}={c=4}{c},
cell{1}{6}={c=3}{c},
cell{10}{2}={font=\bf},
cell{11}{3-8}={font=\bf}} \toprule
& Valid split & & & & Test split \\ \cmidrule[r]{2-5} \cmidrule[l]{6-8}
Method & \FPS & \CER & \EM & \ERone & \CER & \EM & \ERone \\ \midrule
OCR~\cite{MATH25}  & -    & 6.50 & 64.0 & 76.0 & 7.17 & 53.0 & 68.0 \\ \midrule
BTTR $\dagger$     & 8.56 & 6.45 & 60.4 & 72.0 & 6.85 & 53.4 & 65.5 \\
CoMER$\dagger$     & 8.55 & 5.96 & 61.2 & 72.4 & 6.50 & 54.5 & 65.6 \\
ICAL$\dagger$      & 7.95 & 5.43 & 63.4 & 74.5 & 6.03 & 57.7 & 68.2 \\
TAMER$\dagger$     & 6.70 & 5.79 & 61.9 & 73.0 & 6.35 & 55.6 & 66.7 \\
PosFormer$\dagger$ & 5.04 & 5.69 & 62.5 & 73.6 & 6.30 & 56.2 & 67.1 \\ \midrule
GryphOne-MP        & 73.5 & 5.34 & 68.2 & 83.0 & 6.29 & 55.8 & 75.9 \\ \midrule
GryphOne-10        & 73.7 & 4.70 & 70.6 & 84.8 & 5.55 & 59.3 & 78.0 \\
GryphOne-50        & 21.2 & 4.63 & 71.0 & 85.0 & 5.51 & 59.9 & 78.1 \\ \bottomrule
\end{tblr}
\end{table}

\begin{table}[tb]
\centering
\caption{CROHME 2014--2023 results without data augmentation.}
\label{tab:crohme}
\begin{tblr}{
colspec={lcccccccc},
cell{1}{2}={c=8}{c},
cell{2}{2,4,6,8}={c=2}{c},
cell{10}{2-9}={font=\bf}} \toprule
& \ER \\ \cmidrule{2-9}
& 2014 & & 2016 & & 2019 & & 2023 \\ \cmidrule[r]{2-3} \cmidrule[lr]{4-5} \cmidrule[lr]{6-7} \cmidrule[l]{8-9}
Method             & \EM & \ERone & \EM & \ERone & \EM & \ERone & \EM & \ERone \\ \midrule
BTTR$\dagger$      & 57.7 & 67.7 & 52.7 & 63.6 & 54.3 & 64.5 & 53.7 & 64.0 \\
CoMER$\dagger$     & 60.2 & 69.1 & 56.1 & 65.1 & 58.0 & 67.1 & 56.9 & 66.4 \\
ICAL$\dagger$      & 62.4 & 71.2 & 59.0 & 69.1 & 61.0 & 69.4 & 58.3 & 67.4 \\
TAMER$\dagger$     & 59.6 & 69.6 & 55.4 & 66.3 & 58.5 & 68.1 & 57.0 & 66.6 \\
PosFormer$\dagger$ & 59.6 & 69.6 & 58.6 & 68.8 & 58.5 & 68.3 & 56.9 & 66.9 \\ \midrule
NAMER${}^*$        & 60.5 & 75.0 & 60.2 & 73.5 & 61.7 & 75.3 & -    & -    \\ \midrule
GryphOne-50        & 65.2 & 75.9 & 61.4 & 74.3 & 61.8 & 75.7 & 61.2 & 74.8 \\ \bottomrule
\end{tblr}
\end{table}

\subsection{Comparison with Representative Models}

\tabsref{math}{crohme} report the quantitative results.
Baseline methods marked with `$\dagger$' are reimplemented under the same MathWriting training and evaluation pipeline as GryphOne, whereas `$^*$' denotes literature-only scores reported in prior work.
For GryphOne, MP refers to mask-predict~\cite{Mask19} decoding, whereas 10 and 50 indicate the refinement depths $T=10$ and $T=50$, respectively.

Across benchmarks, GryphOne achieves superior ExpRate while maintaining lower CER than AR baselines.
AR decoding cannot revise early errors, whereas NAR decoding depends on one-shot prediction.
In contrast, GryphOne performs iterative refinement that enables global structural correction through stochastic remasking.
The largest gains are observed for near-perfect recognition metrics.

On the MathWriting validation split, EM improves from 63.4 to 71.0 and the $\leq 1$ metric increases from 74.5 to 85.0 compared to ICAL.
On the test split, the $\leq 1$ metric further improves by 9.9 points.
The $T=10$ configuration runs faster than all AR baselines at 73.7 FPS, while preserving over 98\% of the recognition performance of $T=50$.
Although MP outperforms AR baselines, it consistently underperforms GryphOne-10.
Consequently, the diffusion process provides more stable refinement than mask-predict decoding alone.

\begin{table}[tb]
\centering
\caption{Ablation on the MathWriting validation set for diffusion steps $T$.}
\label{tab:abl}
\begin{tblr}{
colspec={cccccccccc},
cell{1}{3,7}={c=4}{c},
cell{4}{9,10}={font=\bf},
cell{6}{3-8}={font=\bf}} \toprule
& & \CER & & & & \SER \\ \cmidrule[r]{3-6} \cmidrule[l]{7-10}
SAT & RMML  & 2    & 5    & 10   & 50   & 2    & 5    & 10   & 50   \\ \midrule
-   & -     & 5.42 & 4.97 & 4.84 & 4.75 & 3.71 & 1.84 & 1.24 & 0.81 \\
-   & \CM   & 5.20 & 4.86 & 4.75 & 4.67 & 2.88 & 1.42 & 0.96 & 0.59 \\ \midrule
\CM & -     & 5.10 & 4.84 & 4.76 & 4.68 & 2.21 & 1.53 & 1.31 & 0.98 \\
\CM & \CM   & 5.01 & 4.78 & 4.70 & 4.63 & 1.89 & 1.33 & 1.08 & 0.84 \\ \bottomrule
\end{tblr}
\end{table}

\subsection{Ablation Study}

\tabref{abl} shows the contributions of SAT and RMML for different total diffusion step counts $T$.
SAT improves CER across all refinement depths, indicating that symbol--modifier alignment helps token-level recognition.
For SER, RMML yields greater improvements at larger $T$, while SAT provides strong early gains at small $T$.
Combining SAT and RMML yields the lowest CER across all diffusion depths while maintaining competitive SER.
These results suggest that SAT and RMML have complementary effects.

\subsection{Diffusion Process Analysis\label{sec:epoch}}

\begin{figure}[tb]
\centering
\subfloat[CER.]{\includegraphics[width=.32\textwidth]{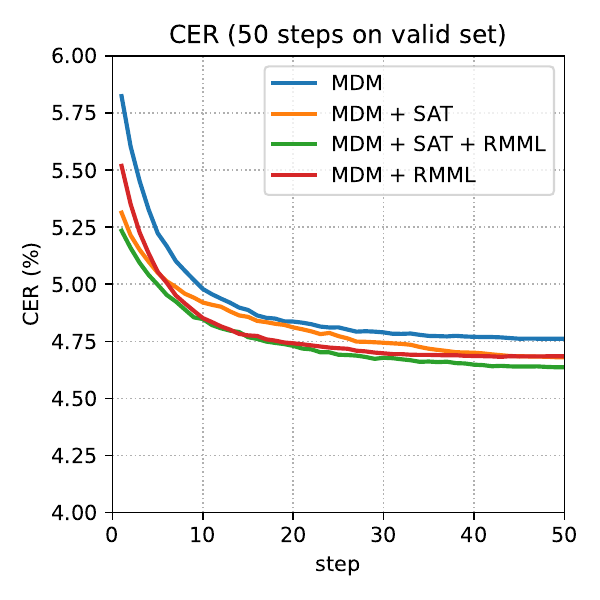}}
\subfloat[EM.]{\includegraphics[width=.32\textwidth]{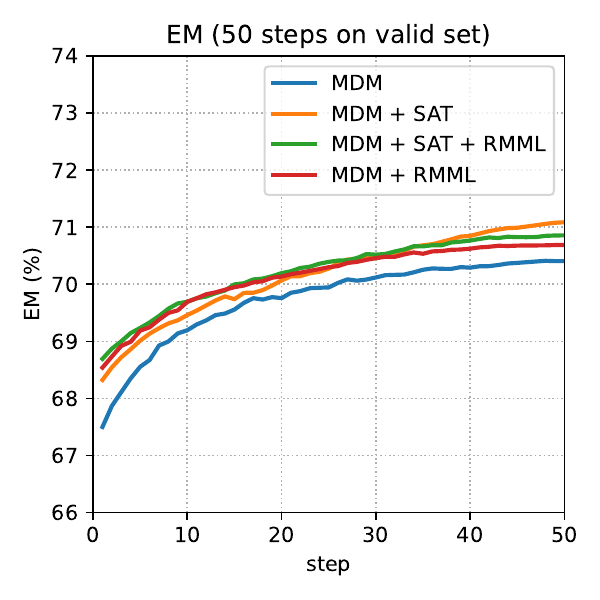}}
\subfloat[SER.]{\includegraphics[width=.32\textwidth]{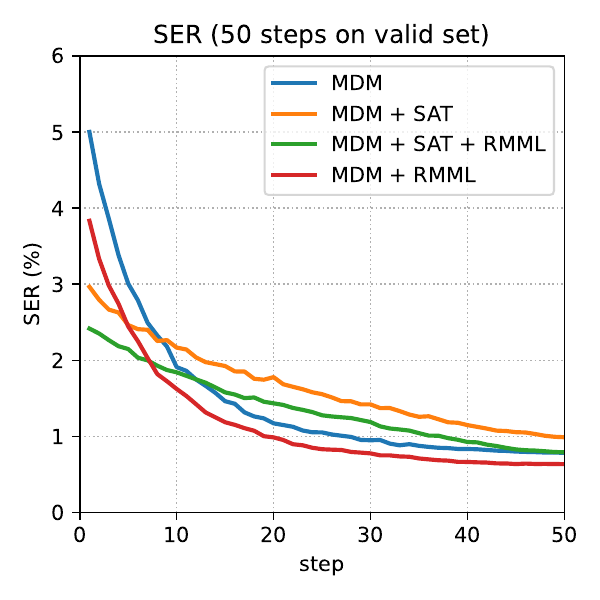}}
\caption{Step-wise error rates on MathWriting ($T=50$).}
\label{fig:process}
\end{figure}

\figref{process} shows the step-wise CER and SER curves on MathWriting during reverse diffusion.
All variants exhibit steady reductions in both rates, with convergence around step 40--50.
This gradual decay reflects the refinement dynamics of reverse diffusion.
Stochastic remasking injects controlled perturbations into the current sequence, preventing premature convergence and enabling revision of predictions as the global context is iteratively updated.
EM (ExpRate) also steadily improves with the refinement steps, as shown in \figref{process}.

SAT consistently lowers CER and SER from the earliest iterations, indicating faster structural alignment.
By operating on symbol--modifier pairs, SAT confines each structural update to a locality-preserving unit.
This accelerates early-stage convergence and stabilizes structural modifications by reducing the propagation of non-local structural dependencies.
RMML further improves stability.

Although SER converges to comparable levels across variants, CER remains consistently lower when SAT is applied.
This suggests that repeated refinement improves not only syntactic validity but also accurate symbol disambiguation in the presence of hierarchical dependencies.
Improvements in both metrics become marginal after approximately 30 steps, suggesting early symbolic and structural stabilization.
Truncated diffusion schedules can retain most of the accuracy while substantially reducing inference cost.

\subsection{Accuracy--Latency Analysis}

\begin{table}[tb]
\centering
\caption{Accuracy--latency trade-off on MathWriting.}
\label{tab:trade}
\begin{tblr}{
colspec={ccccccccc},
cell{1}{2,6}={c=4}{c},
cell{3}{2,6}={font=\bf},
cell{8}{3-5,7-9}={font=\bf}} \toprule
& Valid split & & & & Test split \\ \cmidrule[r]{2-5} \cmidrule[l]{6-9}
$T$ & \FPS & \CER & \EM & \SER & \FPS & \CER & \EM & \SER \\ \midrule
 1 & 164  & 5.23 & 68.7 & 2.42 & 161  & 6.02 & 57.0 & 4.15 \\
 2 & 146  & 5.01 & 69.5 & 1.89 & 144  & 5.82 & 57.9 & 3.18 \\
 5 & 106  & 4.78 & 70.2 & 1.33 & 106  & 5.64 & 58.8 & 2.24 \\
10 & 73.7 & 4.70 & 70.6 & 1.08 & 73.5 & 5.55 & 59.3 & 1.89 \\
20 & 45.2 & 4.66 & 70.7 & 0.94 & 45.4 & 5.53 & 59.6 & 1.74 \\
50 & 21.2 & 4.63 & 71.0 & 0.84 & 21.1 & 5.51 & 59.9 & 1.52 \\ \bottomrule
\end{tblr}
\end{table}

\tabref{trade} reports recognition accuracy and inference speed under different diffusion steps $T$, which determine the accuracy--latency trade-off.
Reducing the diffusion depth maintains near-maximum accuracy while significantly decreasing inference latency compared to the full schedule.
This property highlights a key advantage of diffusion-based recognition.
The refinement depth $T$ serves as an explicit and length-independent control knob for balancing accuracy and latency.
Unlike AR methods, whose latency scales linearly with expression length, GryphOne enables length-independent control of computational cost across diverse applications.

\begin{figure}[tb]
\centering
\subfloat[CER.]{\includegraphics[width=.32\textwidth]{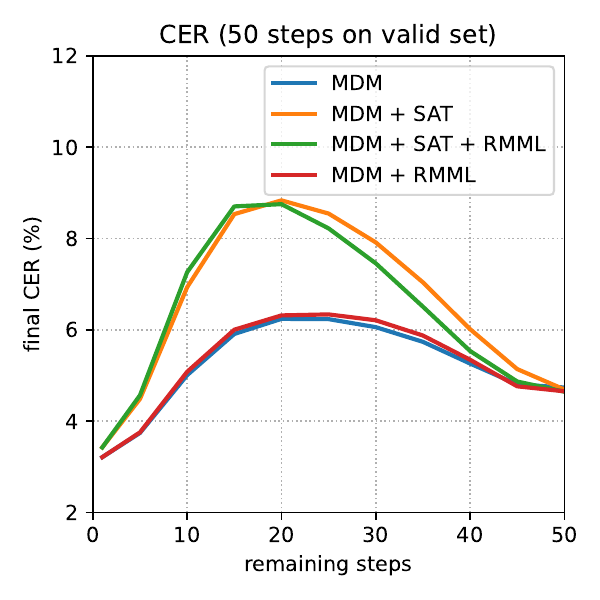}}
\subfloat[EM.]{\includegraphics[width=.32\textwidth]{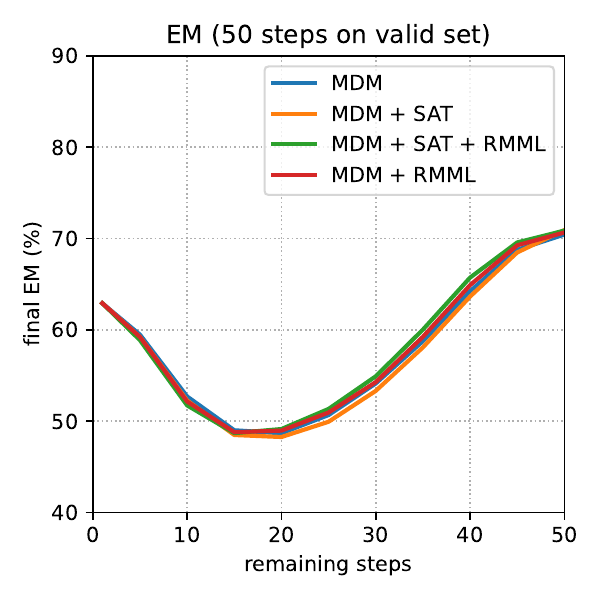}}
\subfloat[SER.]{\includegraphics[width=.32\textwidth]{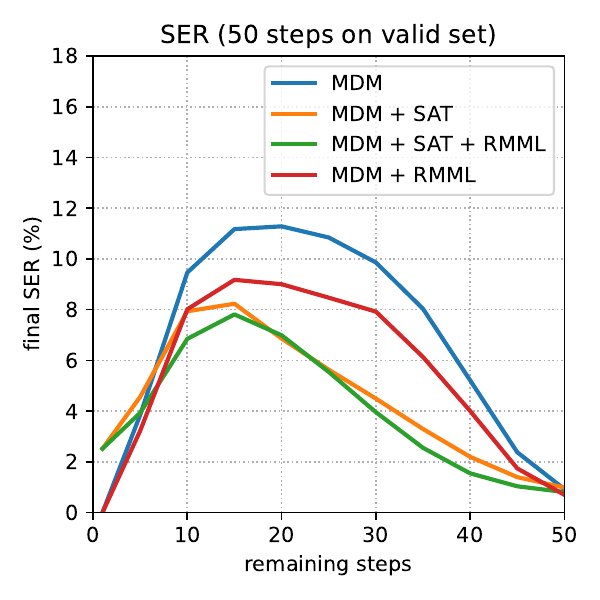}}
\caption{Structural error recovery on MathWriting ($T=50$).}
\label{fig:recovery}
\end{figure}

\subsection{Error Recovery Analysis}

We evaluate structural recovery by removing fraction, square root, integral, and summation operators, while keeping their arguments and braces intact.
Reverse diffusion is initialized at timestep $t$ from the operator-removed sequence, keeping all other symbols unmasked, and run until $t=0$.

\figref{recovery} shows a single extremum around $t\approx20$.
At $t=1$, the sequence remains close to the ground truth except for the removed operators, resulting in inflated accuracy.
At $t=50$, almost all tokens are masked and reconstructed.
The model struggles most to restore partially corrupted sequences at $t\approx20$.
CER is higher with SAT, likely because braces contribute to CER and are removed early during refinement.
In contrast, SER consistently improves with SAT.

\subsection{Diversity Analysis}

\begin{figure}[tb]
\centering
\subfloat[$T=2$.]{\includegraphics[width=.32\textwidth]{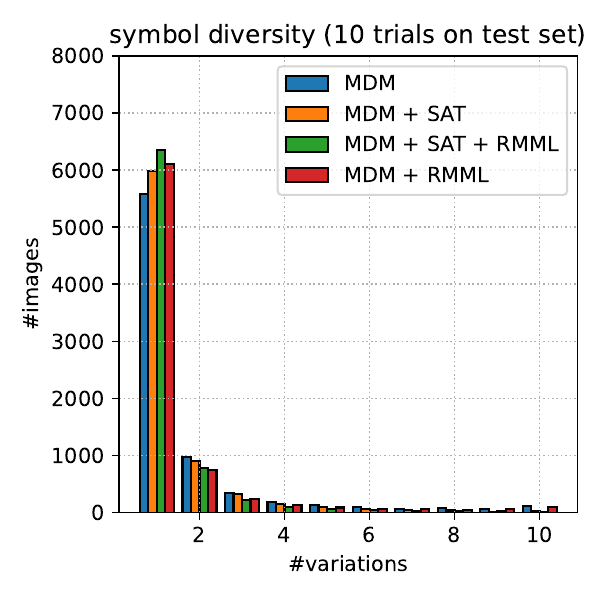}}
\subfloat[$T=10$.]{\includegraphics[width=.32\textwidth]{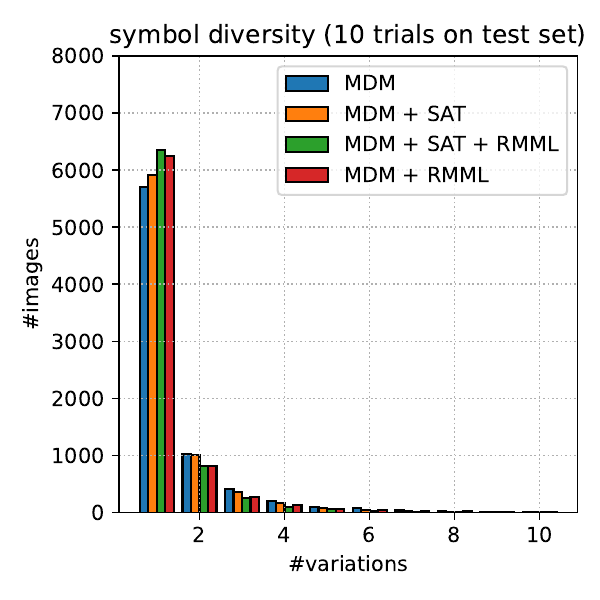}}
\subfloat[$T=50$.]{\includegraphics[width=.32\textwidth]{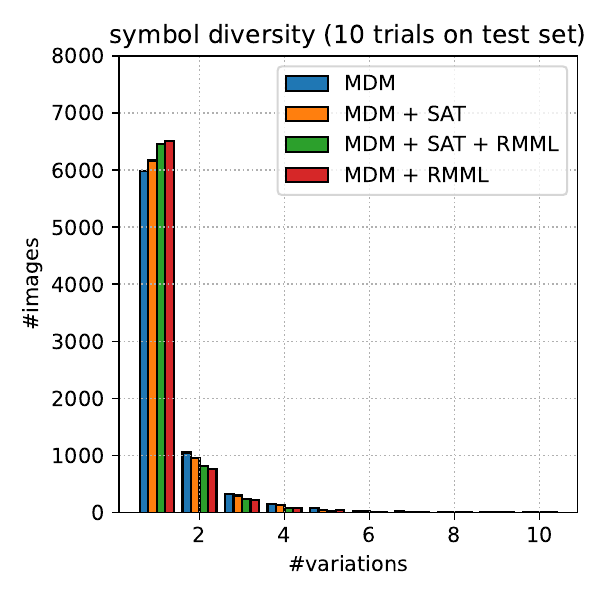}}
\caption{Output diversity analysis on MathWriting.}
\label{fig:hist}
\end{figure}

\figref{hist} shows the distribution of distinct visible symbol sequences obtained from 10 independent decoding runs per input.
Output diversity remains low for most expressions, suggesting stable convergence across stochastic runs.
Both SAT and RMML further reduce variability.

\subsection{Case Study}

\begin{figure}[tb]
\centering
\includegraphics[scale=.3]{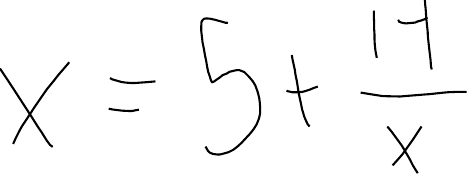} \\
\begin{math}
x=5+\frac{14}{x},
x=5+\frac{\textcolor{red}{17}}{x},
x=5+\frac{\textcolor{red}{4}}{x}.
\end{math}
% from mathwriting_sym_single_mathwriting_50steps.pkl
\caption{Qualitative diversity example for an ambiguous numerator.}
\label{fig:case}
\end{figure}

\figref{case} presents a qualitative example where an ambiguous handwritten fraction leads to multiple plausible outputs.
The alternatives mainly differ in the visually ambiguous numerator, indicating that stochastic decoding reflects genuine visual ambiguity rather than arbitrary output noise.

\section{Conclusion}

GryphOne is a masked diffusion framework that redefines HMER as an iterative refinement process.
GryphOne departs from AR and NAR models by employing stochastic masked reconstruction.
This enables repeated revision of symbols and structural modifiers and mitigates error propagation.
Symbol-aware tokenization enforces locality in structural refinement, while random-masking mutual learning improves refinement stability across iterations.

Extensive evaluation on MathWriting and CROHME demonstrates consistent improvements in complete structural correctness.
Although diffusion introduces additional computational cost, it offers a principled refinement-based alternative to AR inference for symbolic recognition.
Masked diffusion provides an effective framework for vision tasks involving hierarchical dependencies.

\section*{Acknowledgements}

We thank Kenji Aoki and Shinya Shiroshita of Preferred Networks, Inc., for their valuable advice and support.

\bibliographystyle{splncs04}
\bibliography{main}

\end{document}